\documentclass[sigconf]{acmart}

\AtBeginDocument{%
  }

\setcopyright{acmlicensed}
\copyrightyear{2018}
\acmYear{2018}
\acmDOI{XXXXXXX.XXXXXXX}

\acmISBN{978-1-4503-XXXX-X/18/06}

\copyrightyear{2024}
\acmYear{2024}
\setcopyright{rightsretained}
\acmConference[SIGITE '24]{The 25th Annual Conference on Information Technology Education}{October 10--12, 2024}{El Paso, TX, USA}
\acmBooktitle{The 25th Annual Conference on Information Technology Education (SIGITE '24), October 10--12, 2024, El Paso, TX, USA}
\acmDOI{10.1145/3686852.3687083}
\acmISBN{979-8-4007-1106-0/24/10}
\begin{document}

\title{Enhancing classroom teaching with LLMs and RAG}


\author{Elizabeth A Mullins}
\email{emulli@sisd.net}
\affiliation{%
 \institution{Americas High School}
 \city{El Paso}
 \state{Texas}
 \country{USA}
}

\author{Adrian Portillo}
\email{aportillo1@miners.utep.edu}
\affiliation{%
 \institution{Northwest Early College High School}
 \city{El Paso}
 \state{Texas}
 \country{USA}
}
\author{Kristalys Ruiz-Rohena}
\email{kruizrohena@miners.utep.edu}
\affiliation{%
 \institution{University of Texas at El Paso}
 \city{El Paso}
 \state{Texas}
 \country{USA}
}

\author{Aritran Piplai}
\email{apiplai@utep.edu}
\affiliation{%
 \institution{University of Texas at El Paso}
 \city{El Paso}
 \state{Texas}
 \country{USA}
}

\renewcommand{\shortauthors}{Mullins et al.}

\begin{abstract}


Large Language Models have become a valuable source of information for our daily inquiries. However, after training, its data source quickly becomes out-of-date, making RAG a useful tool for providing even more recent or pertinent data. In this work, we investigate how RAG pipelines, with the course materials serving as a data source, might help students in K–12 education. The initial research utilizes Reddit as a data source for up-to-date cybersecurity information. Chunk size is evaluated to determine the optimal amount of context needed to generate accurate answers. After running the experiment for different chunk sizes, answer correctness was evaluated using RAGAs with average answer correctness not exceeding 50 percent for any chunk size. This suggests that Reddit is not a good source to mine for data for questions about cybersecurity threats.
The methodology was successful in evaluating the data source, which has implications for its use to evaluate educational resources for effectiveness.

\end{abstract}




\maketitle

\section{Background and Motivation}




The recent significant advances in the efficacy of large language models (LLMs) have led to a boom in innovative approaches to using LLMs in educational settings and research \cite{akpan2024}. Some uses are elaborate such as the creation of a simulated classroom powered by LLMs acting as teacher, assistant, and student agents managing to demonstrate interaction patterns like real classrooms between agents and other agents as well as between agents and students \cite{zhang2024}. Others examined teacher-led interventions that enhance student interactions with LLMs, aiming to optimize the educational benefits of these tools \cite{kumar2024}.  

\begin{figure}[h]
    \centering
    \includegraphics[width=0.45\textwidth]{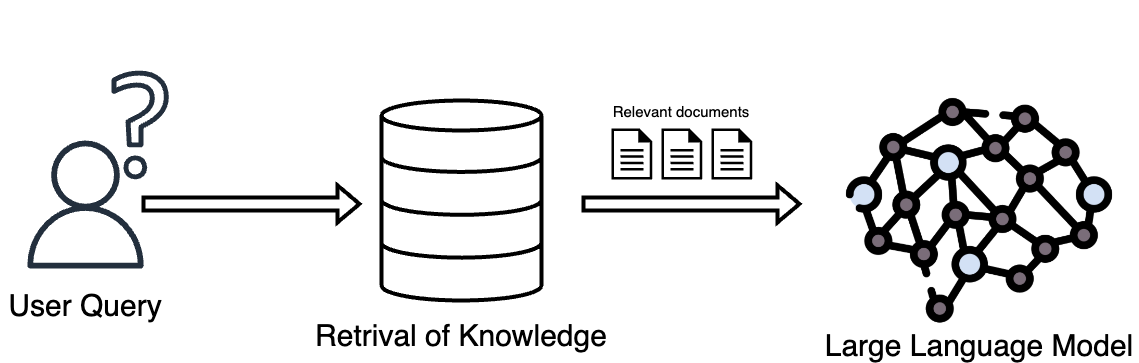}
    \caption{RAG Pipeline}
    \label{fig:label}
\end{figure}

With specialized training, LLMs have shown potential as personalized tutors able to break down problems and prompt students to come up with the answers as well as adjust to the needs of individual learners \cite{chen2023,sonkar2023}. LLMs remain difficult to train with one challenge being ensuring they have updated information to give accurate answers \cite{chen2023}. 

The merging of pre-trained parametric memory (state-of-the-art LLMs) and non-parametric memory, such as a vector database, in retrieval-augmented generation (RAG) models provides a method to give LLMs access to updated information without undergoing new training. These models deliver more specific and factual responses \cite{Lewis2020}. This methodology enhances the potential for LLMs to act as AI tutors \cite{chenxi2024} when connected to the appropriate sources.

\section{Evaluating the utility of LLMs and RAG}
\subsection{Summary of current research}
\begin{figure}[h]
    \centering
    \includegraphics[width=0.45\textwidth]{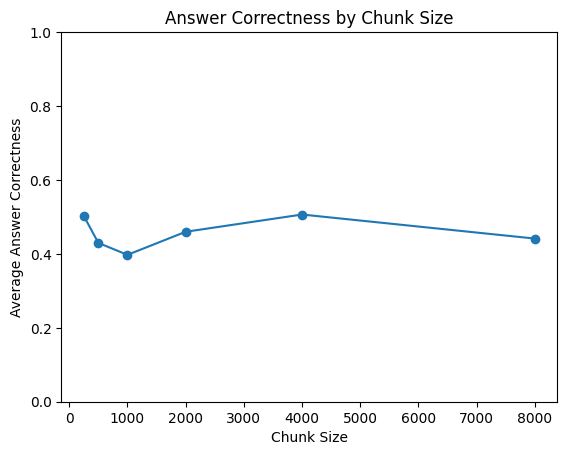}
    \caption{Average answers correctness by chunk sizes.}
    \label{fig:label}
\end{figure}
The research's purpose was to evaluate RAG as a method to use LLMs to provide up-to-date answers to cyber threat questions. In addition, we evaluated the effect of chunk size on the answer correctness of the LLM. We used Llama as our LLM because of its availability. The top 500 posts and the most upvoted answer were scraped from subreddits pertaining to cybersecurity and network security then stored in a Chroma vector database to serve as the data source for the RAG. 30 questions and ground truths were developed from information from CISA about the most recent threats. After feeding the questions through the RAG pipeline, the questions, ground truths, and Llama answers were then fed through RAGAS \cite{ragas} to get an answer correctness score. The process was repeated for chunk sizes of 250, 500, 1000, 2000, 4000, and 8000 characters. The average answer correctness score for each chunk size was close to 50 percent with chunk sizes 250 and 4000 yielding the highest accuracy. 

These results suggest that Reddit is not a good source for a RAG database intended to assist with cybersecurity. Reddit’s format introduces noise such as a variety of postings in a wide range of cybersecurity topics not specific to protecting against current vulnerabilities. In addition, the top posts on Reddit are constantly changing. The first response to a newly posted question can be generated by a bot directing the user to sources of relevant information. Depending on when the data is mined, this could appear multiple times in the context available to the LLM.  

There was little variability in answer correctness by chunk size. Potential effects are thought to be better answers with smaller chunks due to smaller chunks containing more specific context or better answers with larger chunks because larger chunks contain more context. For these specific experiments, there were no generalizable effects. The question bears further examination. With a better aligned data source, chunk size could have a greater impact. Since Reddit did not provide appropriate data for the questions being asked, chunk size would not improve the context provided to the LLM.
\subsection{Future Directions}








This methodology has infinite variations, which is one way to extend it into the classroom. To continue to examine the question of cybersecurity, teachers of cybersecurity content can guide students through using this methodology with alternate sources to mine data for the RAG database. A different database could also lead to noticeable effects on answer correctness with different chunk sizes. This methodology can also be used by students to examine sources for their interests. What sites do they frequent? Can they use this method to mine their favorite sites to provide answers to their questions? To mine data from social media sites to take the trendiness on different topics? This is a methodology that supports student-led/centered research into any topic, making it adaptable to any content and therefore multidisciplinary.

In addition, this model can be used to develop a personalized tutor as demonstrated in the background.  This model allows greater personalization based on student or teacher needs. Other models are designed to be generalized for universal purposes. This model allows for the variations that occur between teachers and students. A teacher who chooses to utilize AI as a teaching assistant, could build the database with their teaching materials, and make the model accessible to their students to query when they need help on a topic. This could be as broad as content for the year or as specific as information needed to complete a project. Changing the data file to create the vector database changes the purpose of the model. For students that do not have a technologically progressive teacher, they can use this model in the same way to create their own tutor.

The model can also be used by teachers to vet the appropriateness of teaching materials or the effectiveness of their assessment questions. Teachers can develop the vector database using their classroom resources such as textbooks and lecture notes or presentations. They can use their assessment questions and answers as the ground truth, then, feed their assessment questions through the RAG pipeline collecting the LLM’s answers to the questions. 
Using RAGAs, teachers can analyze answer correctness of the LLM. If the answer correctness scores are high, this would suggest that the educational materials are appropriate for the learning goals. If the scores are low, this would suggest the materials should be realigned to the essential knowledge and skills.

This work is supported by National Science Foundation Award 2206982



\bibliographystyle{ACM-Reference-Format}
\bibliography{a-bib}


\begin{thebibliography}{8}


\ifx \showCODEN    \undefined \def \showCODEN     #1{\unskip}     \fi
\ifx \showDOI      \undefined \def \showDOI       #1{#1}\fi
\ifx \showISBNx    \undefined \def \showISBNx     #1{\unskip}     \fi
\ifx \showISBNxiii \undefined \def \showISBNxiii  #1{\unskip}     \fi
\ifx \showISSN     \undefined \def \showISSN      #1{\unskip}     \fi
\ifx \showLCCN     \undefined \def \showLCCN      #1{\unskip}     \fi
\ifx \shownote     \undefined \def \shownote      #1{#1}          \fi
\ifx \showarticletitle \undefined \def \showarticletitle #1{#1}   \fi
\ifx \showURL      \undefined \def \showURL       {\relax}        \fi
\providecommand\bibfield[2]{#2}
\providecommand\bibinfo[2]{#2}
\providecommand\natexlab[1]{#1}
\providecommand\showeprint[2][]{arXiv:#2}

\bibitem[Akpan et~al\mbox{.}(2024)]%
        {akpan2024}
\bibfield{author}{\bibinfo{person}{Ikpe~Justice Akpan}, \bibinfo{person}{Yawo~M. Kobara}, \bibinfo{person}{Josiah Owolabi}, \bibinfo{person}{Asuama Akpam}, {and} \bibinfo{person}{Onyebuchi~Felix Offodile}.} \bibinfo{year}{2024}\natexlab{}.
\newblock \bibinfo{title}{An investigation into the scientific landscape of the conversational and generative artificial intelligence, and human-chatbot interaction in education and research}.
\newblock
\newblock
\showeprint[arxiv]{2407.12004}~[cs.CY]
\urldef\tempurl%
\url{https://arxiv.org/abs/2407.12004}
\showURL{%
\tempurl}


\bibitem[Chen et~al\mbox{.}(2023)]%
        {chen2023}
\bibfield{author}{\bibinfo{person}{Yulin Chen}, \bibinfo{person}{Ning Ding}, \bibinfo{person}{Hai-Tao Zheng}, \bibinfo{person}{Zhiyuan Liu}, \bibinfo{person}{Maosong Sun}, {and} \bibinfo{person}{Bowen Zhou}.} \bibinfo{year}{2023}\natexlab{}.
\newblock \bibinfo{title}{Empowering Private Tutoring by Chaining Large Language Models}.
\newblock
\newblock
\showeprint[arxiv]{2309.08112}~[cs.HC]
\urldef\tempurl%
\url{https://arxiv.org/abs/2309.08112}
\showURL{%
\tempurl}


\bibitem[Dong(2024)]%
        {chenxi2024}
\bibfield{author}{\bibinfo{person}{Chenxi Dong}.} \bibinfo{year}{2024}\natexlab{}.
\newblock \bibinfo{title}{How to Build an AI Tutor that Can Adapt to Any Course and Provide Accurate Answers Using Large Language Model and Retrieval-Augmented Generation}.
\newblock
\newblock
\showeprint[arxiv]{2311.17696}~[cs.CL]
\urldef\tempurl%
\url{https://arxiv.org/abs/2311.17696}
\showURL{%
\tempurl}


\bibitem[Es et~al\mbox{.}(2024)]%
        {ragas}
\bibfield{author}{\bibinfo{person}{Shahul Es}, \bibinfo{person}{Jithin James}, \bibinfo{person}{Luis Espinosa~Anke}, {and} \bibinfo{person}{Steven Schockaert}.} \bibinfo{year}{2024}\natexlab{}.
\newblock \showarticletitle{{RAGA}s: Automated Evaluation of Retrieval Augmented Generation}. In \bibinfo{booktitle}{\emph{Proceedings of the 18th Conference of the European Chapter of the Association for Computational Linguistics: System Demonstrations}}. \bibinfo{publisher}{ACL}, \bibinfo{pages}{150--158}.
\newblock


\bibitem[Kumar et~al\mbox{.}(2024)]%
        {kumar2024}
\bibfield{author}{\bibinfo{person}{Harsh Kumar}, \bibinfo{person}{Ilya Musabirov}, \bibinfo{person}{Mohi Reza}, \bibinfo{person}{Jiakai Shi}, \bibinfo{person}{Xinyuan Wang}, \bibinfo{person}{Joseph~Jay Williams}, \bibinfo{person}{Anastasia Kuzminykh}, {and} \bibinfo{person}{Michael Liut}.} \bibinfo{year}{2024}\natexlab{}.
\newblock \bibinfo{title}{Impact of Guidance and Interaction Strategies for LLM Use on Learner Performance and Perception}.
\newblock
\newblock
\showeprint[arxiv]{2310.13712}~[cs.HC]
\urldef\tempurl%
\url{https://arxiv.org/abs/2310.13712}
\showURL{%
\tempurl}


\bibitem[Lewis et~al\mbox{.}(2020)]%
        {Lewis2020}
\bibfield{author}{\bibinfo{person}{Patrick Lewis}, \bibinfo{person}{Ethan Perez}, \bibinfo{person}{Aleksandra Piktus}, \bibinfo{person}{Fabio Petroni}, \bibinfo{person}{Vladimir Karpukhin}, \bibinfo{person}{Naman Goyal}, \bibinfo{person}{Heinrich K\"{u}ttler}, \bibinfo{person}{Mike Lewis}, \bibinfo{person}{Wen-tau Yih}, \bibinfo{person}{Tim Rockt\"{a}schel}, \bibinfo{person}{Sebastian Riedel}, {and} \bibinfo{person}{Douwe Kiela}.} \bibinfo{year}{2020}\natexlab{}.
\newblock \showarticletitle{Retrieval-Augmented Generation for Knowledge-Intensive NLP Tasks}. In \bibinfo{booktitle}{\emph{Advances in Neural Information Processing Systems}}, \bibfield{editor}{\bibinfo{person}{H.~Larochelle}, \bibinfo{person}{M.~Ranzato}, \bibinfo{person}{R.~Hadsell}, \bibinfo{person}{M.F. Balcan}, {and} \bibinfo{person}{H.~Lin}} (Eds.), Vol.~\bibinfo{volume}{33}. \bibinfo{publisher}{Curran Associates, Inc.}, \bibinfo{pages}{9459--9474}.
\newblock
\urldef\tempurl%
\url{https://proceedings.neurips.cc/paper_files/paper/2020/file/6b493230205f780e1bc26945df7481e5-Paper.pdf}
\showURL{%
\tempurl}


\bibitem[Sonkar et~al\mbox{.}(2023)]%
        {sonkar2023}
\bibfield{author}{\bibinfo{person}{Shashank Sonkar}, \bibinfo{person}{Naiming Liu}, \bibinfo{person}{Debshila Mallick}, {and} \bibinfo{person}{Richard Baraniuk}.} \bibinfo{year}{2023}\natexlab{}.
\newblock \showarticletitle{{CLASS}: A Design Framework for Building Intelligent Tutoring Systems Based on Learning Science principles}. In \bibinfo{booktitle}{\emph{Findings of the Association for Computational Linguistics: EMNLP 2023}}, \bibfield{editor}{\bibinfo{person}{Houda Bouamor}, \bibinfo{person}{Juan Pino}, {and} \bibinfo{person}{Kalika Bali}} (Eds.). \bibinfo{publisher}{Association for Computational Linguistics}, \bibinfo{address}{Singapore}, \bibinfo{pages}{1941--1961}.
\newblock
\urldef\tempurl%
\url{https://doi.org/10.18653/v1/2023.findings-emnlp.130}
\showDOI{\tempurl}


\bibitem[Zhang et~al\mbox{.}(2024)]%
        {zhang2024}
\bibfield{author}{\bibinfo{person}{Zheyuan Zhang}, \bibinfo{person}{Daniel Zhang-Li}, \bibinfo{person}{Jifan Yu}, \bibinfo{person}{Linlu Gong}, \bibinfo{person}{Jinchang Zhou}, \bibinfo{person}{Zhiyuan Liu}, \bibinfo{person}{Lei Hou}, {and} \bibinfo{person}{Juanzi Li}.} \bibinfo{year}{2024}\natexlab{}.
\newblock \bibinfo{title}{Simulating Classroom Education with LLM-Empowered Agents}.
\newblock
\newblock
\showeprint[arxiv]{2406.19226}~[cs.CL]
\urldef\tempurl%
\url{https://arxiv.org/abs/2406.19226}
\showURL{%
\tempurl}


\end{thebibliography}

\end{document}